\documentclass[runningheads]{llncs}
\usepackage[T1]{fontenc}
\usepackage{epsfig, soul, pifont}
\usepackage{colortbl}
\usepackage[table]{xcolor}
\usepackage{graphicx}
\usepackage{amsmath}
\usepackage{orcidlink}
\usepackage{amssymb}
\usepackage{amsmath} 
\usepackage{mathtools}
\usepackage{graphicx}
\usepackage{multirow}
\usepackage{subfig}
\usepackage{comment}
\usepackage{authblk}
\usepackage[accsupp]{axessibility}

\usepackage{mathtools}
\usepackage{pgfplotstable}
\usepackage{filecontents,catchfile}
\usepackage{graphicx}
\usepackage{algorithm}% http://ctan.org/pkg/algorithm
\usepackage{algpseudocode}

\usepackage{amsmath}
\usepackage{graphicx}

\begin{document}
\title{STLGRU: Spatio-Temporal Lightweight Graph GRU for Traffic Flow Prediction}

% \thanks{Supported by organization x.}
%
\titlerunning{STLGRU}

\author{Kishor Kumar Bhaumik\inst{1 }\orcidlink{0009-0004-8088-2492} \and
Fahim Faisal Niloy\inst{2}\orcidlink{0009-0003-9466-1458} \and \\  \vspace{-10 pt} Saif Mahmud\inst{3}\orcidlink{0000-0002-5283-0765} \and 
Simon S. Woo\inst{1}\thanks{Corresponding Author}\orcidlink{0000-0002-8983-1542}}

% \author{Kishor Kumar Bhaumik\inst{1} \and
% Fahim Faisal Niloy\inst{2} \\ \and Saif Mahmud\inst{3} \and Simon Woo\inst{1}}
%
\authorrunning{Kishor et al.}
% First names are abbreviated in the running head.
% If there are more than two authors, 'et al.' is used.
%
\institute{Sungkyunkwan University, South Korea \and
University of California, Riverside\and Cornell University\\
% \url{https://dash-lab.github.io/} \and
% ABC Institute, Rupert-Karls-University Heidelberg, Heidelberg, Germany\\
%\email{\{abc,lncs\}@uni-heidelberg.de}}
\email{\{kishor25,swoo\}@g.skku.edu},
\email{fnilo001@ucr.edu},
\email{sm2446@cornell.edu}
}

\maketitle              % typeset the header of the contribution
\vspace{-15pt}
\begin{abstract}
Reliable forecasting of traffic flow requires efficient modeling of traffic data. Indeed, different correlations and influences arise in a dynamic traffic network, making modeling a complicated task. Existing literature has proposed many different methods to capture traffic networks' complex underlying spatial-temporal relations. However, given the heterogeneity of traffic data, consistently capturing both spatial and temporal dependencies presents a significant challenge. Also, as more and more sophisticated methods are being proposed, models are increasingly becoming memory-heavy and, thus, unsuitable for low-powered devices. To this end, we propose \underline{\textbf{S}}patio-\underline{\textbf{T}}emporal \underline{\textbf{L}}ightweight Graph \underline{\textbf{GRU}}, namely \textit{STLGRU}, a novel traffic forecasting model for predicting traffic flow accurately. Specifically, our proposed \textit{STLGRU} can effectively capture dynamic local and global spatial-temporal relations of traffic networks using memory-augmented attention and gating mechanisms in a continuously synchronized manner. Moreover, instead of employing separate temporal and spatial components, we show that our memory module and gated unit can successfully learn the spatial-temporal dependencies with reduced memory usage and fewer parameters. Extensive experimental results on three real-world public traffic datasets demonstrate that our method can not only achieve state-of-the-art performance but also exhibit competitive computational efficiency. Our code is available at \url {https://github.com/Kishor-Bhaumik/STLGRU}

% \url{https://anonymous.4open.science/r/STLGRU}


\end{abstract}

\section{Introduction}
% ,wang2018will,zhang2023rethinking
A traffic network can be represented as a graph, with the locations of the sensors and the connections among them acting as the nodes and edges, respectively. In the same way, flow at a particular junction or node is defined as the total number of people or vehicles passing through that junction at a given time. Specifically, the goal of traffic flow prediction algorithms is to predict the flow of future time steps by exploiting the complex spatialtemporal features of historical traffic data. Indeed, many cities are currently developing Intelligent Traffic Systems (ITS)~\cite{zhang2011data} and predicting traffic flow is a key part of many of these systems' services. In particular, a large amount of collected traffic data have made urban data mining study much easier than ever before, such as traffic flow prediction~\cite{diao2018hybrid}, arrival time estimate~\cite{he2018travel}, traffic speed analysis~\cite{chen2019gated,chen2020multi}, and so on, thanks to the promising advancement of intelligent sensors. 
% ,zhao2017modeling,zhao2018crime
To be more specific, spatio-temporal traffic prediction aims to forecast future traffic trends by analyzing previous spatio-temporal features~\cite{zhao2022multi}. Furthermore, predicting traffic flow has become essential for several downstream applications, such as intelligent route planning~\cite{liebig2017dynamic}, dynamic traffic management~\cite{yang2000simulation}, and location-based services~\cite{lee2011discovering}. However, the efficiency and accuracy of traffic flow prediction algorithms are limited by the high variance in the spatial and temporal dimensions of traffic data. In addition, the observations made at
different locations and time stamps are not independent, but they are rather dynamically correlated. Hence, traffic data has a nonlinear and complex spatial-temporal relationship, and its modeling is critical for designing effective prediction algorithms. 

To address the aforementioned challenges, in this paper, we propose a novel traffic flow prediction model, called Spatio-Temporal Lightweight Graph GRU \textit{(STLGRU)}. Our model takes advantage of graph convolution to model localized spatial relations. We then use an attention mechanism with a memory module to directly model the long-range local and non-local spatio-temporal dependencies. To update the memory, we use a gating mechanism, where our gating strategy records the key local and global spatio-temporal information and forgets the redundant ones when moving to the next time step. In addition, we carefully design our model to be lightweight, as the memory module uses fewer parameters than the existing baselines. Consequently, it can effectively learn long-range dependencies without the need to use multi-scale causal convolution or stacking past time step features. In summary, we make the following contributions:

%Also unlike existing methods, our method does not need to store all previous time steps in a frame for prediction.

\begin{itemize}
    \item We propose \textit{STLGRU}, a novel time series traffic flow prediction model. Our model captures the long-range global and local relationships of a traffic network more accurately by using memory-augmented attention module and gating mechanism.

    \item We carefully design our network to be lightweight by utilizing a memory module with minimal parameters, thus making it suitable for environments constrained by computational resources.

    \item We conduct extensive experiments on three popular traffic prediction benchmark datasets. Our results show that our model not only surpasses other baseline models in performance but also necessitates less memory usage in comparison. 
    
% uses considerably less memory and offers greater computational efficiency. \niloy{we didn't show computation complexity or time analysis. so the word computation efficiency should be discarded. Throughout the paper we should only focus on memory footprint and avoid writing computationally efficient}
    % \st{the memory footprint is lower.}

\end{itemize}

\section{Related Work}

\textbf{Spatio-temporal time series traffic forecasting.} Deep learning has been successfully applied to many tasks, such as image analysis \cite{niloy2021attention,niloy2022cfl}, natural language processing \cite{deb2022variational}, activity recognition \cite{mahmud2020human} etc. Recently, such learning techniques have been quite extensively applied to traffic flow prediction task. Amongst these methods, STGCN~\cite{yu2017spatio} is the first pioneering work to model the traffic network with a fully convolutional structure. In this study, spatio-temporal relationships are effectively captured by including a graph convolution module inside temporal convolution modules. Moreover, DCRNN~\cite{li2017diffusion} introduces diffusion convolution to propagate information in the graph. PM-MemNet~\cite{lee2021learning} learns to match input data to representative patterns with a key-value memory structure. Song et al. proposes STSGCN \cite{song2020spatial}, which captures complex localized spatial-temporal correlation to find the heterogeneities in the spatial-temporal data. STSGCN \cite{song2020spatial} deals with spatial and temporal dimensions individually by utilizing various modules and calculates spatio-temporal attention within a restricted temporal frame.  

In addition, Lin et al. \cite{lin2020self} propose self-attention Conv-LSTM to capture long-range temporal dependencies for general spatio-temporal prediction task. A significant limitation of their approach is its reliance solely on convolution layers, confining their method to spatio-temporal prediction tasks representable by image grids. However, the traffic network has an inherent graph structure that needs to be exploited for reliable prediction. Yuzhou et al.~\cite{chen2021z} tackles this problem by enriching DL architectures with salient time-conditioned topological information of the traffic data. This study introduces the zig-zag persistence concept into time-aware graph convolutional networks.

However, most of the cutting-edge models fail to handle the challenge of being lightweight. RNN-based networks (including LSTM) are widely known to be difficult to train and computationally heavy \cite{yu2017spatio}. For example, Mega-CRN~\cite{jiang2023spatio} proposes Meta-Graph Convolutional Recurrent Network (MegaCRN) by plugging multiple Meta-Graph Learner powered by a MetaNode Bank into the encoder-decoder module. As a consequence, it becomes memory-heavy due to its large number of parameters. STSGCN \cite{song2020spatial} uses a certain length of time window to collect graph structure information and fuse the findings to forecast the following time steps. The computational cost is thus increased by employing repeated shots of graph aggregation. StemGNN \cite{cao2020spectral} introduces a neural network that captures inter-series correlations and temporal dependencies in the spectral domain by aggregating numerous modules in separate blocks while disregarding the model's complexity. To solve the aforementioned issues, we present a simple but effective traffic forecasting model that is computationally cheap, lightweight, and capable of capturing both local and global long-range dependencies in a traffic network.

\textbf{Attention Mechanism.} Because of the high efficiency and versatility in modeling dependencies, attention mechanisms have been extensively used in a variety of domains~\cite{vaswani2017attention,cheng2018neural}. The basic principle behind attention mechanisms is to concentrate on the most relevant features of the input data~\cite{cheng2018neural}. Recently, researchers used attention processes to graph-structured data to model spatial correlations for graph classification~\cite{velivckovic2017graph}.  We expand the attention method to synchronize spatial and temporal dependencies while sequentially predicting traffic data.

\begin{figure*}[ht!]
\centering
\includegraphics[width=1.0 \textwidth]{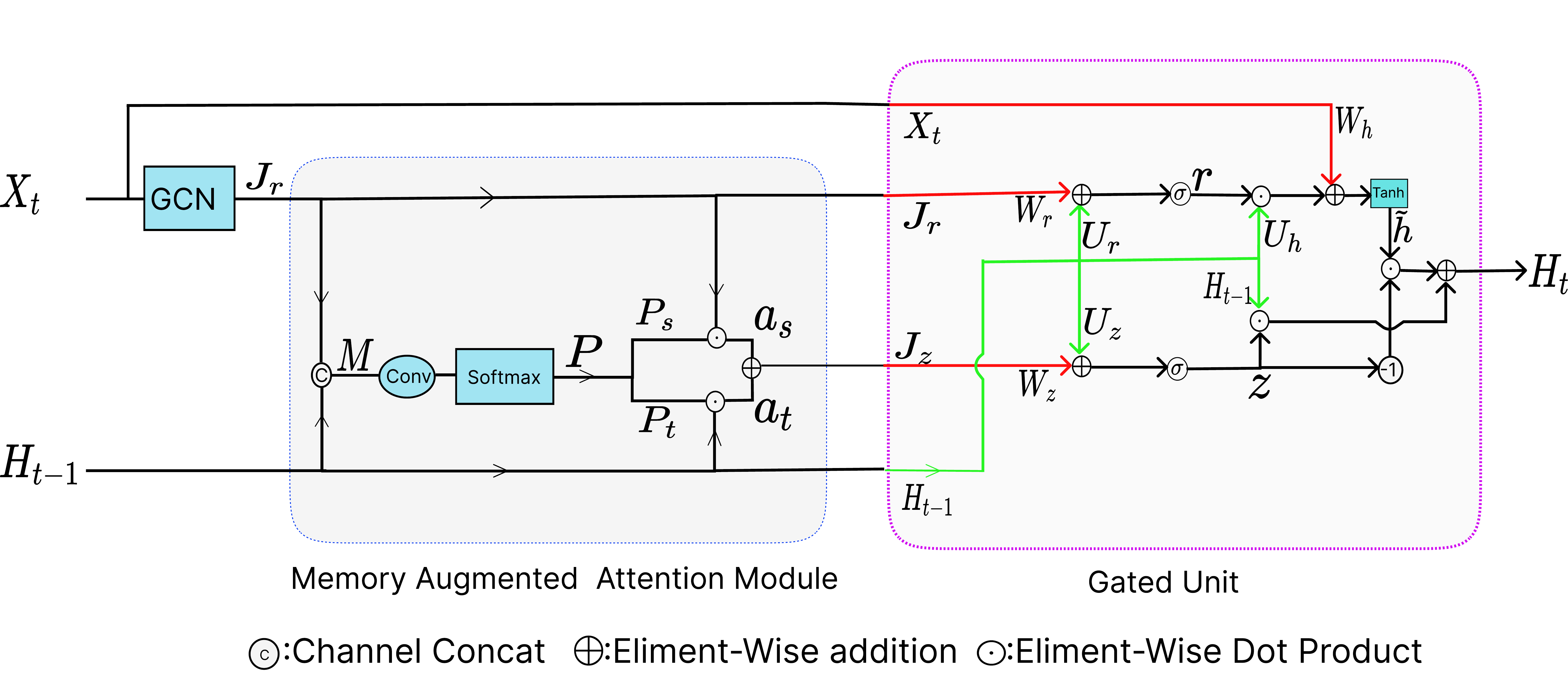}
\caption{ Overall architecture of STLGRU designed for multivariate traffic forecasting. Our model consists of a memory-augmented attention module and a gated unit, which capture the long-range local and global dependencies. It takes input from a single time step with an initial hidden state and outputs a hidden state for the next time step.}
\label{fig:stlgru}
\end{figure*}
%Although STGCN employed basic temporal and spatial blocks, its computational efficiency is greater than ours owing to the stacked network. 
\vspace{-25pt}
\section{Proposed Model}

\subsection{Preliminaries and Problem Definition} A graph $\mathcal{G}=(\mathcal{V},\mathcal{E})$ models the traffic topological network, where $\mathcal{V}$ represents nodes and $\mathcal{E}$ signifies edges. An edge $e_{ij} \in \mathcal{E}$ connects nodes $v_i$ and $v_j$, where each node has junctional features (e.g. inflow, outflow). Then, we define the spatio-temporal traffic data forecasting problem using a mapping function $f_\theta$, where it takes the historical series $\langle X_{(t-T+1)}, X_{(t-T+2)}, \ldots, X_{t}\rangle$. And, it predicts the future series $\langle X_{(t+1)}, X_{(t+2)}, \ldots, X_{\left(t+T^{\prime}\right)} \rangle$, where $T$ is the length of the historical series and $T^{\prime}$ is the length of the target forecast series, and,  $X_i\in \mathbb{R}^{N \times C}$, where $N$ is the number of nodes and $C$ is the number of information channels (speed, flows, etc.).Thus, the time series forecasting model can be defined as follows:
{\small
\begin{align*}
\langle X_{(t-T+1)}, X_{(t-T+2)}, \ldots, X_{t}\rangle \xrightarrow{f_{\theta}}\langle X_{(t+1)}, X_{(t+2)}, \ldots, X_{\left(t+T^{\prime}\right)} \rangle
\end{align*}
}
%\subsection{Our STLGRU Model} 

% Fig. \ref{overflow} depicts the architecture of our proposed \textit{STLGRU}. 

\subsection{Graph Convolution } Here, we first define the graph convolution, where the initial input matrix is denoted as $X \in \mathbb{R}^{N \times T \times C}$. As our focus is solely on traffic flow, $C$ is consequently set to 1, resulting $X \in \mathbb{R}^{N \times T \times 1}$. And, we take $X_{t^{\prime}} \in \mathbb{R}^{N \times 1}$ as input from a single time step $t$, where $t \in T$, and pass it through a convolutional layer $\xi_\theta$ to transform the input feature into high-dimensional space $C^{\prime}$ to increase the representation power of the network as follows:

\begin{equation}
\label{eqn:eqn2}
\begin{gathered}
X_t=\xi_{\theta}\left(X_{t^{\prime}}\right); \theta \in \mathbb{R}^{1 \times C^{\prime} } 
\end{gathered}
\end{equation}

\noindent Then, $X_t\in R^{N \times C^{\prime}}$ is used as an input to the original network at time step $t$.
\vspace{-14pt}
\begin{figure}[ht!]
\centering
\includegraphics[width=80mm, scale=0.5]{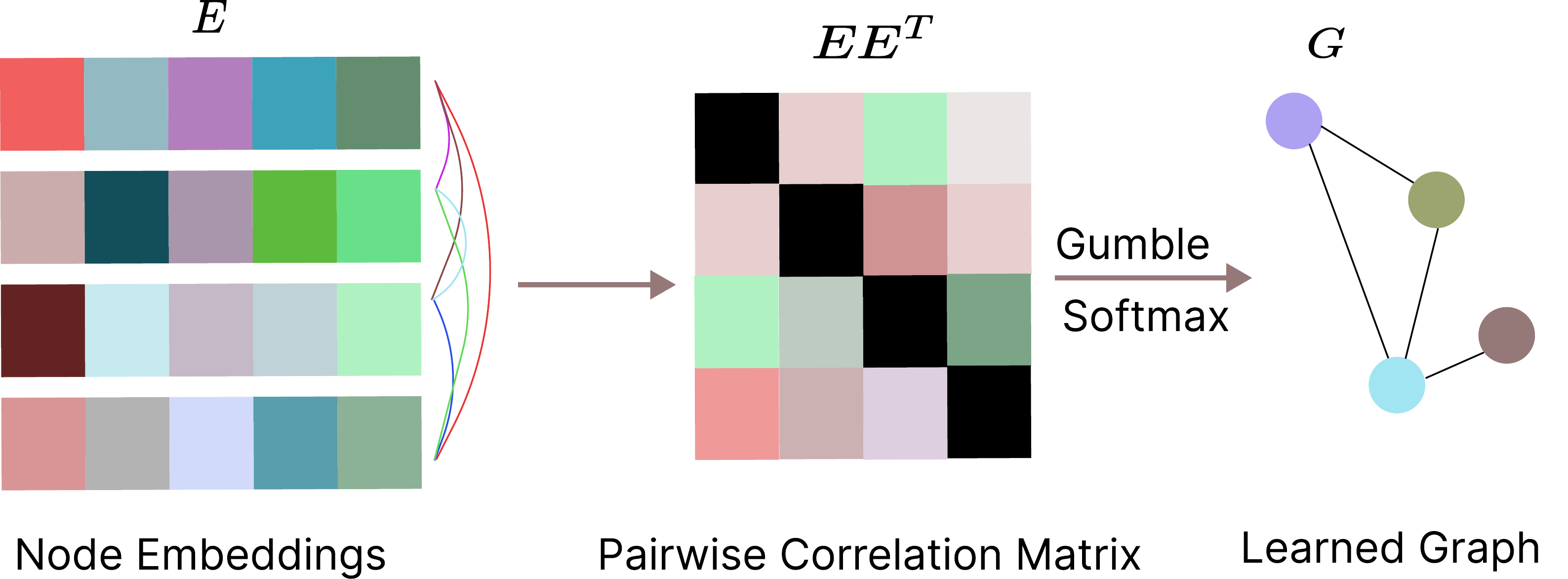}
\caption{ Graph generation from learnable node embeddings $E$ }
\label{gcn}
\end{figure}

As shown in Fig.~\ref{gcn}, let $E \in \mathbb{R}^{N \times d}$ be the learned node embedding matrix, where $d$ represents the embedding dimension. In addition, $\Omega$ represents the probability matrix, and each $\Omega_{ij} \in \Omega$  corresponds to the probability of preserving the edge between time series $i$ and $j$, respectively. This relationship is formally expressed as follows:
\vspace{-5pt}
\begin{equation}
\Omega = E{E}^T
\end{equation}
\vspace{-10pt}

In particular, we use the Gumbel softmax method~\cite{jang2016categorical} to obtain the final sparse adjacency matrix $A \in R^ {N \times N}$ to effectively assure a sufficient amount of sparsity in the graph structure. And, let $\sigma$ and $\tau$ be the activation function and the temperature variable, respectively. Then, we can define sparse adjacent matrix $A$ as follows:

\begin{equation}
\label{eqn:eqngumble}
\begin{gathered}
A = \sigma((log(\Omega_{ij}/(1-\Omega_{ij})+(n^1_{ij}-n^2_{ij})/\tau) \\
s.t.\;n^1_{ij}, n^2_{ij} \sim Gumbel(0,1)
\end{gathered}
\end{equation}

Eq.~\eqref{eqn:eqngumble} implements Gumbel Softmax for our task, where $A_{i,j}= 1 $ with probability $\Omega_{i,j}$ and $A_{i,j}= 0$ with the remaining probability. In particular, Gumbel Softmax maintains the same probability distribution as the normal Softmax, ensuring statistical consistency in generating the trainable probability matrix for the graph forecasting network. Next, let $I$ be an identity matrix and $D$ be a diagonal degree matrix satisfying $D_{ii}= \Sigma_j A_{ij}$. Then, the specific operation of graph convolution network (GCN) with the learnable weight $W \in R^{C^{\prime} \times C^{\prime}}$ can be expressed as follows:
\vspace{-5pt}
\begin{gather}
\label{eqn:eqn14}
G C N\left(X_t\right)= W(I + D^{-\frac{1}{2}}AD^{-\frac{1}{2}})X_t \in \mathbb{R}^{N \times C^{\prime}}
\end{gather}

\subsection{Memory-Augmented Attention (MAA) Module} 

As discussed before, many state-of-the-art models struggle with maintaining a lightweight design. For instance, models proposed by Jiang et al. [1] and Yu et al. [2] have learned spatio-temporal relations by combining GCN and GRU modules and they further stack these fused modules multiple times. However, when stacking multiple layers to capture long-term dependencies in traffic data, they encounter a significant increase in memory usage during inference. To mitigate this issue, we introduce a memory-augmented attention (MAA) mechanism by continuously synchronizing relevant features in both spatial and temporal data in each timestep. In particular, Figure~\ref{fig:stlgru} illustrates the MAA module's structure, where it combines the graph convolution output $J_r \in \mathbb{R}^{N \times C^{\prime}} $ with the randomly initialized hidden input $H_{t-1} \in \mathbb{R}^{N \times C^{\prime}}$ through concatenation, and pass it through a convolutional function as follows:
\vspace{-7pt}
\begin{align}
 &J_r=G C N\left(X_t\right) ,\\
&M = J_r \oplus {H_t} , \\
&P=\psi_{w}\left(M\right),
\end{align}

\noindent where $\psi$ is a 1D convolutional function with parameter $w \in\mathbb{R}^{ C^{\prime} \times C^{\prime}} $, and $M$  and $P$ both have the same dimension of  $\mathbb{R}^{2N \times C^{\prime}} $. We use the softmax specified by $P_{u,v}$ to calculate the attention score for both spatial characteristics from the GCN output and temporal features from the hidden input in the following way: 

\begin{align}
&P_{u, v}=\frac{\exp P_{u,v}}{\sum_{v=1}^N \exp P_{u,v}}, u,v \in\{1,2, \ldots, N\}.
\end{align}

After the above step, $P \in \mathbb{R}^{2N \times C^{\prime}} $  is divided into $P_s$ and $P_t$, which have the same size, $ \left(P_s,P_t\right) \in \mathbb{R}^ {N \times C^{\prime} }$. Next, we element-wise multiply $P_s$ and $P_t$ with $J_r$ and $H_{t-1}$ respectively, as follows:
\vspace{-10pt}
\begin{align}
&a_s = P_s \odot J_r,\\
&a_t = P_t \odot {H_{t-1}},
\end{align}

\noindent where $\odot$ represents the Hadamard product. Rather than exclusively representing spatial context, $a_s$ also includes temporal information for a specific timestamp, while $a_t$ serves a similar dual role, encompassing both spatial and temporal context. We then add these two context vectors, finally producing $J_z$ as follows: 
\vspace{-10pt}
\begin{align}
&J_z = a_s+a_t; J_z \in \mathbb{R}^{N \times C^{\prime}}
\end{align}

% \begin{figure}[!ht]
% \centering
% \includegraphics[width=70mm]{GRUcell.pdf}
% \caption{ Vanilla Gated Recurrent Unit (GRU) }

% \label{gru}
% \end{figure}

\subsection{Memory Updating} Prior traffic forecasting models~\cite{yu2017spatio,fang2019gstnet,guo2019attention} often use graph and temporal convolution independently, overlooking the heterogeneities within spatial-temporal data. To tackle this problem, our approach involves a continuously synchronized gating mechanism to update the hidden state $H_t$, allowing MAA to capture long-range dependencies across both spatial and temporal domains effectively. The update process is defined as follows:
\begin{align}
g &=\sigma\left(W_z \cdot J_z+U_z \cdot H_{t-1}\right),\\
r &=\sigma\left(W_r \cdot J_r+U_r \cdot H_{t-1}\right),\\
\tilde{h} &=\tanh \left(W_h \cdot X_t +r * U_h \cdot H_{t-1} \right),\\
H_t &=g * H_{(t-1)}+(1-g) * \tilde{h},
\end{align}

\noindent where $ \left (W,U \right) \in \mathbb{R}^ { C^{\prime} \times C^{\prime}} $ are the learnable parameters and $\sigma$ is the sigmoid function. Compared with the original memory cell in the GRU~\cite{chung2014empirical} that is updated only by current input $X_t$ and previous hidden state $H_{t-1}$, Our proposed memory cell updates based on the original input $X_t$, graph convolution $J_r$, aggregated context vector $J_z$, and the previous hidden state $H_{t-1}$, which effectively captures both local and global spatio-temporal dependencies in real-time.
% Therefore, we argue that $H_{t-1}$ can contain global past spatiotemporal information.

On the other hand, similar to the standard GRU mechanism, we use the final output at the last time step, denoted as $H_T\in R^{N \times C^{\prime}}$, and process it through two fully connected layers for prediction as follows:

\begin{align}
    \hat{\mathcal{Y}}=\operatorname{Re}LU\left(H_T W_1+b_1\right) \cdot W_2+b_2, 
    \label{function}
\end{align}

\noindent where $\hat{\mathcal{Y}}\in \mathbb{R}^{N \times T^{\prime}}$ denotes the prediction of the overall network, and $W_1 \in$ $\mathbb{R}^{ C^{\prime} \times C^{\prime}}, b_1 \in \mathbb{R}^{C^{\prime}}, W_2 \in \mathbb{R}^{C^{\prime} \times T^{\prime}}$, and $b_2 \in \mathbb{R}^{T^{\prime}}$ are learnable parameters. Finally, to train the model, we use the loss function as follows:
\vspace{-5pt}
\begin{align}
\mathcal{L}(\theta)=\left\|\widetilde{\mathcal{Y}}-\hat{\mathcal{Y}}\right\|_2^2 %=\left,\|\widetilde{\mathcal{Y}}-f_\theta\right\|_2^2,
\label{final}
\end{align}
where  $\widetilde{\mathcal{Y}}$ denotes the ground truth and $\hat{\mathcal{Y}}$ denotes the prediction of the model, respectively. %\textbf{The detailed algorithm of our proposed method is provided in the supplementary section}.
\textbf{The detail of our training algorithm is provided in the supplementary material.}

\section{Experimental Results and Analysis } 

\noindent \textbf{Datasets.} We perform experiments on three publicly available popular benchmark traffic datasets, which are PeMSD4, PeMSD7, and PeMSD8 from California Transportation Agencies~\cite{chen2001freeway}. In these datasets, each vertex on the graph represents a sensor node to collect the traffic flow data and the flow data is aggregated to 5 minutes. Thus, each hour has 12 data points in the flow data. We apply zero-mean normalization for preprocessing these datasets. 

% The summary statistics of the key elements of the datasets are shown in Table~\ref{tab:dataset}.
% \input{table_dataset.tex}

% \begin{figure*}
% \centering
% %\includegraphics[width=85mm]{sensor_dist.pdf}
% \includegraphics[width=1.65\columnwidth]{sensor_dist.pdf}
% \caption{ Sensor Distribution of three traffic datasets, where the dots are the traffic-sensor locations.} 
% \label{sensordist}
% \end{figure*}

\noindent \textbf{Baselines.}
We compare our proposed \textit{STLGRU} against the following popular as well as SoTA baseline models on spatio-temporal prediction task: 1) Spatial-temporal synchronous modeling mechanism (STSGCN~\cite{song2020spatial}), 2) Spectral Temporal Graph Neural Network for time series forecasting (StemGNN~\cite{cao2020spectral}), 3) Time Zigzags at Graph Convolutional Networks (Z-GCNETs~\cite{chen2021z}), 4) Graph-Wavenet (GW-Net~\cite{wu2019graph}),  5) Pattern Matching Memory Networks (PM-MemNet~\cite{lee2021learning}), and 6) Meta-Graph Convolutional Recurrent Network (Mega-CRN\cite{jiang2023spatio}).  We use default settings for each baseline when performing comparisons.
\begin{center}
   \begin{table*}[!t]
\centering
\caption{The overall performance of STLGRU and baseline methods.}
\resizebox{1.0\linewidth}{!}{
\begin{tabular}{cc||ccc|ccc|ccc}
\hline
                          &                                               &                                        & 15 min                                 &                                        &                                        & 30 min                                 &                                        &                                        & 60 min                                 &                                        \\ \cline{2-11} 
\multirow{-2}{*}{Datasets} & Model                                         & MAE                                    & RMSE                                   & MAPE                                   & MAE                                    & RMSE                                   & MAPE                                   & MAE                                    & RMSE                                   & MAPE                                   \\ \hline

                          & STSGCN                                        & 19.41                                  & 30.69                                  & 14.82                                  & 21.83                                  & 31.33                                  & 15.54                                  & 23.19                                  & 33.65                                  & 16.90                                  \\
                          & StemGNN                                       & 20.24                                  & 28.15                                  & 13.03                                  & 20.68                                  & 30.88                                  & 14.21                                  & 22.92                                  & 33.74                                  & 15.65                                  \\

                          & Z-GCNETs                                      & 19.50                                  &  28.61                            & 12.78                                  & 23.21                                  & 30.09                                  & \underline{13.12}                            & 29.24                                  & 32.95                                  & 16.14                                  \\
             PeMSD4     & GW-Net                                        & \underline{18.15}                            & 25.24                                 & 13.27                                  & 22.12                                  & 30.62                                  & 16.28                                  & \underline{21.85}                            & 33.70                                  & 17.29                                  \\
                          & PM-MemNet                                     & 18.95                                  & 30.16                                  & 13.79                                  & 20.01                                  & 31.47                                  & 14.17                                  & 26.85                                  & 32.14                                  & 17.21                                  \\
                          & Mega-CRN                                      & 19.25                                  & \underline{24.88}                                 & \underline{12.72}                            & \underline{19.60}                            &\underline{25.96}                            & 13.84                                  & 22.82                                  & \underline{26.33}                            & \underline{14.87}                            \\ \hline
\multirow{-11}{*} & \cellcolor[HTML]{FFFFFF}\textbf{STLGRU(Ours)} & \cellcolor[HTML]{FFFFFF}\textbf{17.59} & \cellcolor[HTML]{FFFFFF}\textbf{23.24} & \cellcolor[HTML]{FFFFFF}\textbf{11.02} & \cellcolor[HTML]{FFFFFF}\textbf{18.73} & \cellcolor[HTML]{FFFFFF}\textbf{24.61} & \cellcolor[HTML]{FFFFFF}\textbf{12.85} & \cellcolor[HTML]{FFFFFF}\textbf{21.05} & \cellcolor[HTML]{FFFFFF}\textbf{25.41} & \cellcolor[HTML]{FFFFFF}\textbf{13.87} \\ \hline
 
                          & STSGCN                                        & 16.17                                  & 23.15                                  & 16.51                                  & 22.19                                  & 34.87                                  & 19.88                                   & 24.26                                  & 39.03                                  & 20.21                                  \\
                          & StemGNN                                       & 15.77                                  & 22.68                                  & 13.97                                  & 22.38                                  & 33.69                                  & 18.99                                   & 24.54                                  & 34.41                                  & 19.45                                   \\

                       & Z-GCNETs                                      & 15.64                                  & 25.19                                  & 15.47                                  & 23.78                                  & 33.64                                  & 19.05                                  & 26.12                                  & 34.78                                  & 23.47                                  \\
        PeMSD7          & GW-Net                                        & 18.74                                  & 26.14                                  & 16.58                                  & 23.64                                  & 34.82                                      & 24.65                                  & 24.15                                  & 34.12                                            & 29.02                                  \\
                          & PM-MemNet                                     & 15.25                                  & 24.14                                  & 15.17                           & \underline{21.12}                            & 34.41                                  & 19.97                                   & 25.39                                  & \underline{ 33.50}                            & 21.29                                 \\
                          & Mega-CRN                                      & \underline{14.23}                            & \underline{21.05}                            & \underline{13.11}                                 & 22.86                                  & \underline{33.19}                            & \underline{18.40}                             & \underline{23.55}                            & 33.54                                  & \underline{19.29}                             \\ \hline
                          & \cellcolor[HTML]{FFFFFF}\textbf{STLGRU(Ours)} & \cellcolor[HTML]{FFFFFF}\textbf{13.79} & \cellcolor[HTML]{FFFFFF}\textbf{19.12} & \cellcolor[HTML]{FFFFFF}\textbf{12.31} & \cellcolor[HTML]{FFFFFF}\textbf{20.89} & \cellcolor[HTML]{FFFFFF}\textbf{31.45} & \cellcolor[HTML]{FFFFFF}\textbf{15.56}  & \cellcolor[HTML]{FFFFFF}\textbf{23.06} & \cellcolor[HTML]{FFFFFF}\textbf{32.19} & \cellcolor[HTML]{FFFFFF}\textbf{19.12}  \\ \hline

                          & STSGCN                                        & 15.97                                  & 23.14                                  & 14.79                                  & 16.45                                  & 24.78                                  & 18.47                                  & 19.13                                  & 29.80                                  & 18.96                                  \\
                          & StemGNN                                       & 15.83                                  & 24.93                                  & 10.26                                  & 15.95                                  & 23.88                                  & 19.98                                  & 24.10                                  & 28.13                                  & 23.79                                  \\

                          & Z-GCNETs                                      & 15.76                                  & 25.11                                  & 10.01                                 & \underline{15.64}                            & 23.29                                  & 16.67                                  & \underline{17.55}                            & 29.67                                  & 19.19                                  \\
        PeMSD8             & GW-Net                                        & 14.95                                  & 24.92                                  & 12.79                                  & 15.92                                  & 24.99                                         & 18.97                                  & 17.69                                  & 28.92                                          & 22.67                                  \\
                          & PM-MemNet                                     & 14.10                                  & \underline{22.15}                            & 10.41                                  & 16.65                                  & 24.17                                  & \underline{13.77}                             & 19.13                                  & 28.16                                  & \underline{16.68}                                  \\
                          & Mega-CRN                                      & \underline{14.07}                            & 22.53                                  & \underline{9.54}                             & 16.10                                  & \underline{22.42}                            & 17.97                                   & 18.12                                  & \underline{27.29}                            & 21.05                            \\ \hline
                          & \cellcolor[HTML]{FFFFFF}\textbf{STLGRU(Ours)} & \cellcolor[HTML]{FFFFFF}\textbf{13.93} & \cellcolor[HTML]{FFFFFF}\textbf{20.94} & \cellcolor[HTML]{FFFFFF}\textbf{8.84}  & \cellcolor[HTML]{FFFFFF}\textbf{15.03} & \cellcolor[HTML]{FFFFFF}\textbf{22.18} & \cellcolor[HTML]{FFFFFF}\textbf{12.64}  & \cellcolor[HTML]{FFFFFF}\textbf{16.83} & \cellcolor[HTML]{FFFFFF}\textbf{26.35} & \cellcolor[HTML]{FFFFFF}\textbf{14.74} \\ \hline
\end{tabular}
}
\label{Tab:table1}
\end{table*}
 
\end{center}

\vspace{-28pt}
\noindent \textbf{Evaluation Metrics.} We apply three widely used metrics to evaluate the performance of our model, (1) Mean Absolute Error (MAE), (2) Mean Absolute Percentage Error (MAPE), and (3) Root Mean Squared Error (RMSE).

\noindent \textbf{Implementation Details.} We divide all the datasets with a ratio 6:2:2 into training, testing, and validation sets, respectively.  We use Adam optimizer with a learning rate of 0.001 and set 16 as the batch size. We conduct experiments with our model using non-overlapping time windows in the time series data. The entire experiments are run on a single GPU (Nvidia TITAN RTX). If the test scores of a baseline are unknown for a dataset, we run their publicly available code based on their suggested settings to obtain the results.
%\vspace{-10pt}

\noindent \textbf{Results.} Table \ref{Tab:table1} compares the performance of our model to the baseline models in 15, 30 and 60 minutes traffic forecasting, respectively. As shown in Table \ref{Tab:table1}, our model outperforms all of the baseline models in both long and short-term forecasting. StemGNN, Z-GCNET, STSGCN, GW-Net, and PM-MemNet stack multiple layers of spatio-temporal modules by optimizing a probabilistic graph model. Our proposed method demonstrates improvements over the comparative models, achieving an average increase of 2.7\%, 3.1\%, and 2.3\% in MAE, RMSE, and MAPE, respectively. Mega-CRN, which utilizes trainable adjacency matrices to understand node relationships and employs an encoder-decoder structure to manage traffic data heterogeneity, is also surpassed by our STLGRU model. Overall, STLGRU demonstrates superior performance, exhibiting average improvements of  2.9\%, 3.1\%, and 2.6\% in MAE, RMSE, and MAPE, respectively.

% \vspace{-4pt}

\begin{center}
\begin{table*}[]
\centering
\caption{In-depth comparison of different model efficiency on PeMSD4. We show that STLGRU achieves high memory efficiency with less computational power and parameters in all three datasets. The second best is shown with underline (See Supp. for more results with additional datasets).}
\begin{tabular}{ccccc}
\hline
\multirow{1}{*}{Model }          
                       & Memory (MB)   & FLOPs            & Parameters       \\ \hline
STSGCN                  & \underline{1028}        & 282.24G         & \underline{550.48K}      \\
GW-Net                  & 1031         & \underline{189.16G}        & 610.25K             \\
StemGNN                & 1220         & 378.98G         & 1.64M                  \\
Z-GCNETs               & 1473         & 389.49G         & 1.08M        \\
Mega-CRN               & 1409         & 311.97G         & 669.14K          \\
PM-MemNet              & 1052         & 421.49G         & 1.34M           \\ \hline
\textbf{STLGRU (Ours)}  & \textbf{990} & \textbf{77.93G} & \textbf{348.54K}  \\ \hline
\end{tabular}
\label{tab:memory_table}
%\vspace{-20pt}
\end{table*}
\vspace{-30pt}

\end{center}

Furthermore, Table \ref{tab:memory_table} presents the maximum memory footprint, computational complexity, and the number of parameters of the baseline models on PeMSD4 dataset. Because we use the same model for each dataset, we present the experimental results with one dataset. Results with additional datasets are provided in Suppl. To compute a model's GPU memory usage during inference, we use the Linux command line ``gpustat'' with a minibatch size of 1 and with no gradients. We can observe that \textit{STLGRU} requires the least memory during inference than baseline models. It also has the least computation complexity and number of parameters. In Table~\ref{tab:memory_table}, we present the memory footprint, computational complexity, and the number of parameters used to train \textit{STLGRU}. Our model stands out, as it demands the least memory, with fewer parameters, and ours exhibits reduced computational complexity. This efficiency positions our model as an ideal choice for integration into real-world, low-powered devices.

% \begin{figure} [!ht]
% \centering
% %\includegraphics[width=50mm,scale=0.5]{method.eps}
% \includegraphics[width=75mm, scale=0.5]{forcast_plot.pdf}
% \caption{ Visualization of the predicted traffic flow.}
% \label{fig:forecasting}
% \end{figure}

\section{Ablation Study} 

We verify the effectiveness of \textit{STLGRU} with additional ablation experiments. We dissect our model and focus on two main components: the Gumbel softmax and the memory augmented attention (MAA). As illustrated in Table~\ref{tab:ablation}, the absence of MAA leads to a remarkable decline in performance. The role of Gumbel softmax is pivotal in ensuring optimal sparsity within the graph. When we substitute Gumbel softmax with only the learnable embedding matrix, there is a noticeable decline in our model's performance. However, this is not surprising, given that irrelevant connections can reduce the model's ability to capture the dynamic interrelations between nodes accurately.  

% \vspace{-5pt}
\begin{figure*}[!h]
\centering
\includegraphics[width=1 \textwidth]{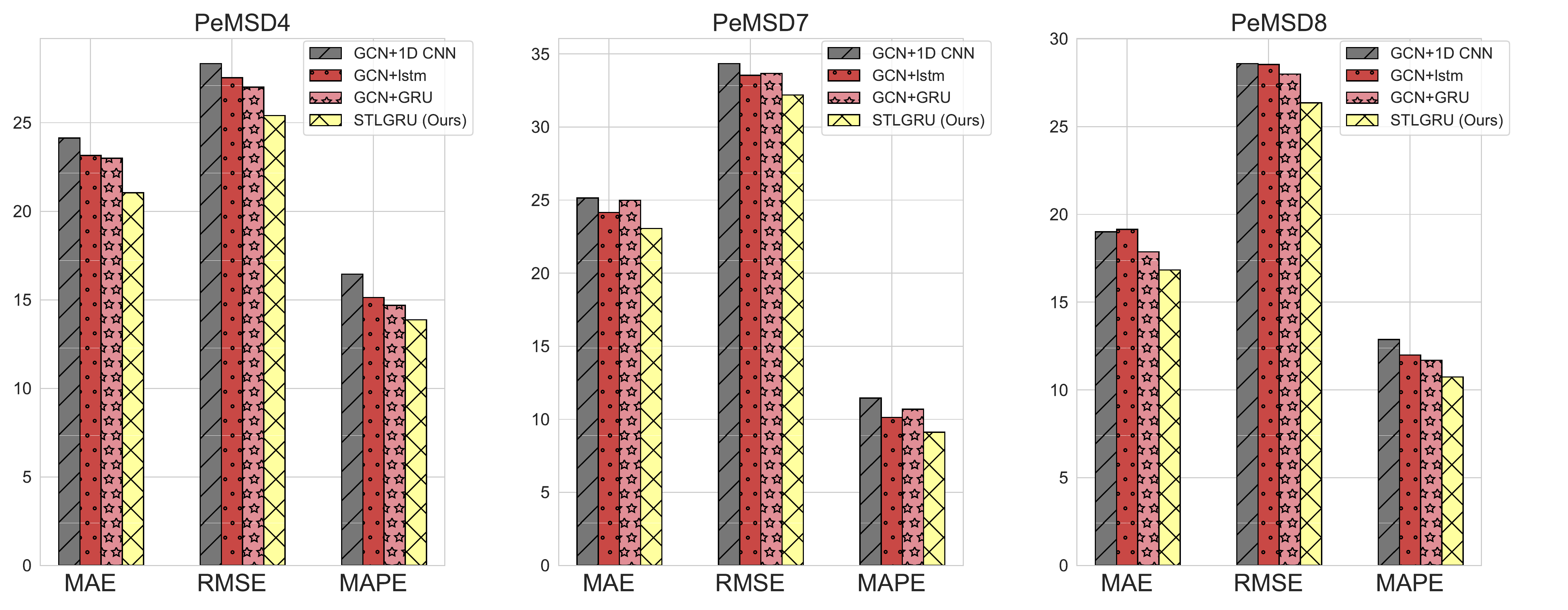}
\caption{  Performance comparison of spatio-temporal models and \textit{STLGRU} with different settings. MAE, RMSE and MAPE of 1-hour forecasting on three datasets are plotted.}
\label{fig:visualization}
\end{figure*}

% \vspace{-10pt}

\begin{comment}
    
\begin{table}[!h]
\centering
\caption{MAE scores based on different output feature size in PEMS dataset.}
\begin{tabular}{c|c|c|cc}
\hline
\begin{tabular}[c]{@{}c@{}}Output  features\end{tabular} & PEMS04                     & PEMS07                    & PEMS08                     \\ \hline
8                                                           & 21.97          & 23.09        & 16.95         \\ 
16                                                           & 22.01         & 23.14         & 16.91         \\ 
32                                                                 & 21.95         & 23.15         & 16.89          \\ 
64                                                        & \textbf{21.90} & \textbf{23.06} & \textbf{16.83} \\ 
128                                                              & 22.03          & 23.18         & 17.14         \\ \hline
\end{tabular}
\\[5pt]
\label{tab:ablation}
\end{table}
\end{comment}

\begin{table}[h!]
\centering
\caption {Ablation study for the effectiveness of the memory augmented attention (MAA) and gumble softmax module used in our method.}
\vspace{10pt}
\label{tab:ablation}
\begin{tabular}{c|c|c}
\hline
 Gumble Softmax & MAA & Error Score (MAE) \\ \hline
$\times$ & $\times$  & 23.12 \\
$\times$ & \checkmark & 21.74 \\
\checkmark & $\times$  & 19.83 \\
\checkmark & \checkmark & \textbf{16.83} \\ \hline
\end{tabular}
\end{table}

Afterwards, as demonstrated in Figure~\ref{fig:visualization}, we compare our model against traditional spatio-temporal configurations. Specifically, we evaluate (1) Graph convolution for capturing spatial knowledge and 1D CNN to capture temporal dependencies, (2) Graph convolution for capturing spatial knowledge and LSTM to capture temporal dependencies, (3) Graph convolution for capturing spatial knowledge and vanilla GRU to capture temporal dependencies. From our observations in Figure~\ref{fig:visualization}, it is evident that \textit{STLGRU} consistently outperforms other methods significantly. We thus argue that memory-augmented attention can capture more fine-grained spatio-temporal patterns and trace the crucial interdependencies among the road network.
\vspace{-10pt}

\section{Conclusion}

In this work, we introduce \textit{STLGRU}, a uniquely lightweight and efficient model for traffic flow prediction task. Our model incorporates a memory module enhanced with attention mechanism, capable of synchronizing spatial correlations within node networks and long-term temporal patterns in a continuous manner. Our experimental results showcase its superior performance across three benchmark traffic prediction datasets while maintaining a significantly reduced computational overhead compared to baseline models. For future work, we plan to adapt \textit{STLGRU} for other spatial-temporal forecasting challenges, and explore how to model spatio-temporal dependencies when long-term data is scarce.

\noindent \small{\textbf{Acknowledgements}. This work was partly supported by Institute for Information \& communication Technology Planning \& evaluation (IITP) grants funded by the Korean government MSIT: (No. 2022-0-01199, Graduate School of Convergence Security at Sungkyunkwan University), (No. 2022-0-01045, Self-directed Multi-Modal Intelligence for solving unknown, open domain problems), (No. 2022-0-00688, AI Platform to Fully Adapt and Reflect Privacy-Policy Changes), (No. 2021-0-02068, Artificial Intelligence Innovation Hub), (No. 2019-0-00421, AI Graduate School Support Program at Sungkyunkwan University), and (No. RS-2023-00230337, Advanced and Proactive AI Platform Research and Development Against Malicious Deepfakes). Lastly, this work was supported by Korea Internet \& Security Agency (KISA) grant funded by the Korea government (PIPC) (No.RS-2023-00231200, Development of personal video information privacy protection technology capable of AI learning in an autonomous driving environment). 
}\normalsize
%\section{Acknowledgeent} This work was supported by Institute of Information \& communications Technology Planning \& Evaluation (IITP) grant by the Korea government (MSIT) (No.RS-2023-00230337,Advanced and Proactive AI Platform Research and Development Against Malicious Deepfakes).

\bibliographystyle{splncs04}
\bibliography{refs,strings}

\clearpage
\begin{center}
\Large\textbf{Supplementary Material for\\STLGRU: Spatio-temporal Lightweight Graph GRU for Traffic Flow Prediction}
\end{center}
\vspace{0.1cm} % Adjust vertical spacing as needed

\setcounter{section}{0}
\section{Proposed Training Algorithm for STLGRU}
We describe the training algorithm of our proposed \textit{STLGRU} in Algorithm~\ref{stlgru}.
\vspace{-10pt}

\begin{algorithm}
  \caption{\textit{STLGRU}}\label{stlgru}
  \begin{algorithmic}[1]
  \item \textbf{Input:} $X = \langle X_{(t-T+1)}, X_{(t-T+2)}, \ldots, X_{t}\rangle; \quad X \in \mathbb{R}^{N \times T \times 1}$
  \item \textbf{Output:} $ \widetilde{Y} = \langle  X_{(t+1)}, X_{(t+2)}, \ldots, X_{\left(t+T^{\prime}\right)}\rangle; \quad \widetilde{Y} \in \mathbb{R}^{N \times T \times 1}$
  \item \textbf{Parameters:} \texttt{Randomly initialize $\Theta$ and hidden state $H_{t-1}$}

      \For{\texttt{all} $T$}
        \State $X_{t^{\prime}} \gets X[:,t,:]; \quad   X_{t^{\prime}} \in \mathbb{R}^{N \times 1} $ 
        \State $X_t=\xi_{\theta}\left(X_{t^{\prime}}\right) ;  \quad X_t\in R^{N \times C^{\prime}}$ \Comment{followed by eq. ~\ref{eqn:eqn2}}
        \State $H_{t-1} = {STLGRU}_{\Theta} \left( X_t,H_{t-1} \right)$
      \EndFor
      \State $H_t= H_{t-1}$
      \State $\hat{Y} = OutputLayer(H_t)$  \Comment{followed by eq. ~\ref{function}}
      \State calculate loss $L$ using eq. ~\ref{final}
      \State \textbf{return} $\hat{Y} $
    % \EndProcedure
  \end{algorithmic}
\end{algorithm}
\vspace{-10pt}
\section{Memory consumption}
In this section, we compare our proposed model with existing baselines using PeMSD7 and PeMSD4 datasets.
\vspace{-20pt}
\begin{table*}[]
\centering
\caption{In-depth comparison of different model efficiency. We show that \textit{STLGRU} achieves high memory efficiency with less computational power and parameters in all three datasets. The second best is shown with underline.}
\resizebox{1.0\linewidth}{!}{ 
\begin{tabular}{c||ccc|ccc}
\hline
\multirow{2}{*}{Model} &  &  PeMSD7 &                   &       & PeMSD8          &                \\ 
                         & Memory (MB)    & FlOPs            & Parameter        & Memory (MB)   & FlOPs           & Parameter       \\ \hline

STSGCN                   & 1420          & \underline{384.27G}         & 895.73K          & \underline{920}          & \underline{88.55G}          & \underline{98.7K}           \\
StemGNN                       & 1816          & 589.94G          & 1.87M            & 1111         & 271.51G         & 1.22M           \\
Z-GCNETs             & 1753          & 442.629G         & 1.45M            & 1314         & 298.26G         & 987.25K         \\
Mega-CRN            & 1638          & 421.85G          & 1.12M            & 1312         & 245.68G         & 889.79K         \\
GW-Net                   & 1920          & 497.64G          & \underline{827K}             & 1037         & 144.91G         & 247.63K         \\
PM-MemNet                       & \underline{1267}          & 512.73G          & 1.64M            & 934        & 437.61G         & 1.07M           \\ \hline
\textbf{STLGRU(Ours)}  & \textbf{1328} & \textbf{295.54G} & \textbf{634.89K} & \textbf{893} & \textbf{52.15G} & \textbf{79.82K} \\ \hline
\end{tabular}
}

\label{tab:memory_table_sup}
\end{table*}
\vspace{-30pt}

\section{Prediction visualization}
\begin{figure} [!ht]
\centering
\includegraphics[width=75mm, scale=0.3]{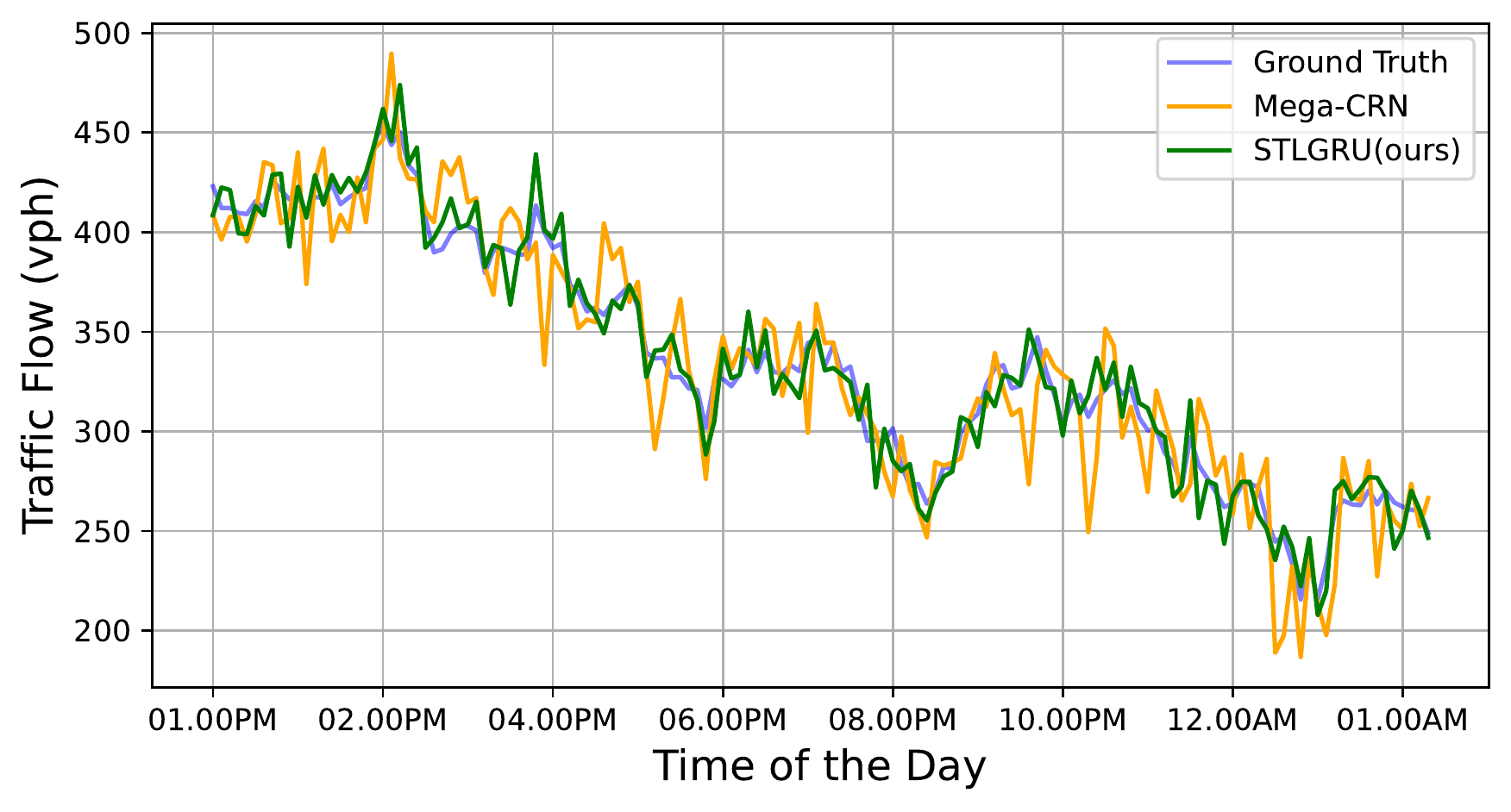}
\caption{ Visualization of the predicted traffic flow.}
\label{fig:forecasting}
\end{figure}

We visually plot the time series alongside its ground truth in Figure~\ref{fig:forecasting}. This comparative visualization underscores STLGRU’s superior predictive capabilities compared to the baseline Mega-CRN.

\vspace{-25pt}
\section{Dataset discription }

The summary statistics of the key elements of the datasets are shown in Table~\ref{tab:dataset}.
\vspace{-20pt}

\begin{table}[ht!]
\centering
\caption{Dataset Statistics.}
\begin{tabular}{ccccc}
\hline
Datasets & Nodes & Edges & Timesteps & Periods                 \\ \hline
PeMSD4   & 307   & 340   & 16,992    & 2018/01/01 - 2018/02/28 \\
PeMSD7   & 883   & 866   & 28,224    & 2016/07/01 - 2016/08/31 \\ 
PeMSD8   & 170   & 277   & 17,856    & 2016/07/01 - 2016/08/31 \\ \hline
\end{tabular}
% \end{table}
\label{tab:dataset}
\end{table}

\vspace{-25pt}
\begin{figure*}
\centering
\includegraphics[width=.7\columnwidth]{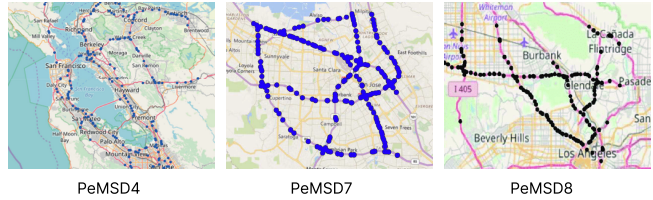}
\caption{ Sensor Distribution of three traffic datasets, where the dots are the traffic-sensor locations.} 
\label{sensordist}
\end{figure*}
\vspace{-20pt}

\section{Baseline discription }
We compare our proposed \textit{STLGRU} against the following baseline models on spatial-temporal prediction task. 

\begin{itemize}

\item  [\ding{71}] STSGCN: STSGCN captures the complex localized spatial-temporal correlations through a spatial-temporal synchronous modeling mechanism.
\item [\ding{71}] StemGNN: StemGNN combines Graph Fourier Transform (GFT) which models inter-series correlations and Discrete Fourier Transform (DFT) which models temporal dependencies in an end-to-end framework.

\item [\ding{71}] Z-GCNETs: Z-GCNETs proposes to enhance DL architectures
with the most salient time-conditioned topological information of the data and introduce the concept of zigzag persistence into time-aware graph convolutional networks.
\item [\ding{71}] GW-Net: Graph-Wavenet developed a novel adaptive dependency matrix and learned it through node embedding which can precisely capture the hidden spatial dependency in the data.
\item [\ding{71}] PM-MemNet: PM-MemNet learns to match input data to representative patterns with a key-value memory structure.
\item [\ding{71}] Mega-CRN: Meta-Graph Convolutional Recurrent Network (MegaCRN) uses the Meta-Graph Learner incorporating a MetaNode Bank into GCRN encoder-decoder.
\end{itemize}

\section{Evaluation metrics}
Mean Absolute Error (MAE), Mean Absolute Percentage Error (MAPE) and Root Mean Squared Error (RMSE) are derived as follows:
\begin{align}
    &M A E=\frac{1}{n} \sum_{t=1}^{t=n}\left|y^{\prime}-y\right| \\
    &MAPE =\frac{1}{n} \sum_{t=1}^{t=n} \frac{\left|y^{\prime}-y\right|}{y} * 100 \% \\
    &R M S E=\sqrt{\frac{1}{n} \sum_{t=1}^{t=n}\left(y^{\prime}-y\right)^2}
\end{align}

\end{document}